\documentclass[letterpaper, 10 pt, conference]{ieeeconf}  

\IEEEoverridecommandlockouts                              

\usepackage{bm,amsmath,amsfonts,amssymb,dsfont} 
\usepackage{mathrsfs} 
\usepackage{multirow} 
\usepackage{graphicx} 
\graphicspath{{figures/}}

\usepackage{hyperref}
\hypersetup{
    colorlinks=true,
    linkcolor=blue,
    filecolor=magenta,
    urlcolor=blue,
}

\newcommand{\NewR}{\ensuremath{\mathds{R}}}
\newcommand{\mat}[1]{{\ensuremath{{\mathbf{#1}}}}}
\newcommand*\diff{\mathop{}\!\mathrm{d}}

\DeclareMathOperator{\Tr}{Tr}

\usepackage{siunitx} 

\DeclareMathOperator*{\argmax}{argmax} 

\title{\LARGE \bf
Precise Object Placement with Pose Distance Estimations for Different Objects and Grippers
}

\author{
\begin{small}
Kilian Kleeberger$^{1}$,
Jonathan Schnitzler$^{1}$,
Muhammad Usman Khalid$^{1}$,
Richard Bormann$^{1}$,
Werner Kraus$^{1}$,
and Marco F. Huber$^{1,2}$
\end{small}
\thanks{$^{1}$Fraunhofer Institute for Manufacturing Engineering and Automation IPA,
        Nobelstra{\ss}e~12, 70569 Stuttgart, Germany
        {\tt\small kilian.kleeberger@ipa.fraunhofer.de}}%
\thanks{$^{2}$Institute of Industrial Manufacturing and Management IFF, University of Stuttgart,
        Allmandring~35, 70569 Stuttgart, Germany
        {\tt\small marco.huber@ieee.org}}%
}

\begin{document}

\maketitle
\thispagestyle{empty}
\pagestyle{empty}

\begin{abstract}
This paper introduces a novel approach for the grasping and precise placement of various known rigid objects using multiple grippers within highly cluttered scenes. Using a single depth image of the scene, our method estimates multiple 6D object poses together with an object class, a pose distance for object pose estimation, and a pose distance from a target pose for object placement for each automatically obtained grasp pose with a single forward pass of a neural network.

By incorporating model knowledge into the system, our approach has higher success rates for grasping than state-of-the-art model-free approaches. Furthermore, our method chooses grasps that result in significantly more precise object placements than prior model-based work.
\end{abstract}





\section{Introduction}
\begin{figure}[t]
\centering
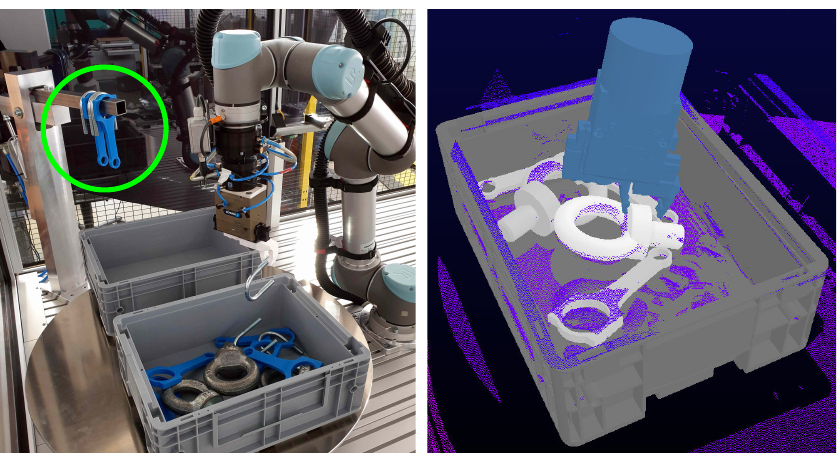
\caption{(left) Real-world robot cell for picking chaotically stored objects of different categories from bins. Objects can be placed on the bar in the background (see green circle).
(right) 3D point cloud (colored) with pose estimates of different object types (light gray) of our approach including the bin (gray) on real-world data without ICP refinement. The gripper (blue) is visualized at the top-ranked grasp pose.
}
\label{fig:result_real_world_data}
\vspace{-3mm}
\end{figure}

For robots to reliably grasp and manipulate objects in undefined poses, they have to perceive their environment by means of sensors and plan corresponding actions accordingly.
%
%
In this work, we focus on robotic \mbox{bin-picking}, where multiple rigid objects of different types are stored chaotically in a bin and the robot has to pick the objects and place them at a given target pose as exemplarily visualized in Fig.~\ref{fig:result_real_world_data}.
%

\begin{figure}[tb]
\centering
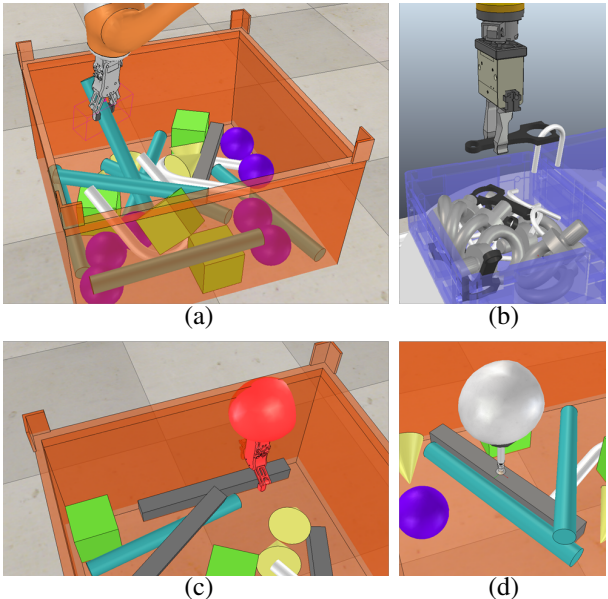
\caption{Failure cases of robotic \mbox{bin-picking} tasks:
(a) The robot picks an object (IPACylinder) that moves relative to the gripper during lifting due to overlaps with other objects. The object cannot be placed precisely anymore. (b) The robot picks an object (IPAConnectingRod) that is entangled with another object (IPAUBolt). This can result in a dropping of objects during placement or collisions with the environment at the defined target pose for placement.
(c) An object (Bar) cannot be picked due to collisions with the bin.
(d) An object (Bar) cannot be picked with a parallel jaw gripper due to collisions with other objects in the scene. Using a suction gripper allows picking these objects.
}
\label{fig:Motivation}
\vspace{-3mm}
\end{figure}

The task is challenging due to a high amount of clutter and occlusion in the scene, various types of
objects that have to be differentiated, object symmetries that result in pose ambiguities, varying lighting conditions, and missing,
incorrect, and noisy depth information from the real-world sensor. Furthermore, collisions with the bin and other objects in the scene have to be avoided.
During grasping, the targeted object can easily move relative to the gripper under the weight of any objects that are on top of it as shown in Fig.~\ref{fig:Motivation}~(a) or when the grasp is too far off from the center of mass of the object. Additionally, some object geometries can potentially cause entanglements as illustrated in Fig.~\ref{fig:Motivation}~(b). This can lead to collisions when placing the picked object at the defined target pose or other objects dropping during placement.

Often, the bins cannot be emptied completely, due to no collision-free reachability of objects~\cite{FPS_GPC_1,FPS_GPC_2}. An object at the border of a bin may not be graspable with a parallel jaw gripper because one finger collides with the bin wall as exemplarily visualized in Fig.~\ref{fig:Motivation}~(c). The same holds, when, e.g., bars or cylinders lie very close next to each other as illustrated in Fig.~\ref{fig:Motivation}~(d). In these cases, the candidates can be picked with an eccentric suction gripper where the suction cup can still be placed on top of the objects.
Furthermore, different object geometries are generally challenging to handle with a single gripper type.



In this paper, we tackle these challenges by providing a multi-gripper approach which executes grasping trials in simulation and transfers that gained experience to the real world to be able to select highly robust grasps on different object types.
Our approach jointly solves the 6D object pose estimation (OPE), object classification, and grasps quality prediction tasks while using common gripper types such as parallel jaw and suction grippers.
%
Our method decides autonomously which object with which gripper including grasp pose (or gripping point) is best suited for execution.
Furthermore, it works for all possible kinds of object symmetries
for both OPE and grasping with placement.

Current robotic \mbox{bin-picking} solutions~\cite{ISRAVISIONAG.2021,Spenrath.2013,Spenrath.2017,paper_TEM}
require manual parameterization and tuning of the object localization and
gripping algorithms
until a satisfactory system performance is reached~\cite{ReviewArticleSpringer_KBK,Kraft.2014,FPS_GPC_2}.
%
%
Through the use of object models (CAD or previously scanned~\cite{Paper_RMB} models),
our approach can
adapt itself autonomously to new objects
without any human intervention.
%
Our approach is entirely trained on synthetic
images and annotations and can, therefore, be easily applied to new objects
by triggering a new data generation and neural network training.

To the best of our knowledge, we are the first to use pose distance estimations for object placement based on grasping and placement trials in simulation for different objects and grippers simultaneously.
%
We provide challenging mixed bins datasets that can also be used by other researchers for benchmarking.
All datasets and videos of real-world experiments are available at~\url{https://www.bin-picking.ai/en/dataset.html}.
%
Although implemented for \mbox{bin-picking}, our approach can be used for other pick-and-place tasks such as shelf picking, depalletizing, conveyor belt picking, etc.,
while especially the latter is an attractive application due to the high speed and real-time capability of our method.

In summary, the main contributions of this work are:
\begin{itemize}
    \item Novel approach \mbox{PQ-Net++} for grasping and placing objects of different categories using a multi-gripper policy.
    \item Estimation of the pose distance for object pose estimation to select the most promising pose estimates.
    \item Estimation of the pose distance for object placement to favor grasps resulting in precise placements.
    \item Two novel challenging benchmark datasets and performance evaluation on these datasets.
\end{itemize}




\section{Related Work}
\label{sec:RelatedWork}
Approaches to robotic grasping and manipulation can be categorized along multiple criteria~\cite{ReviewArticleSpringer_KBK}. In the following, we differentiate between model-free and model-based approaches based on whether model knowledge of the object to pick is used or not.

\subsection{Model-free Approaches}
Model-free grasping poses a dominant direction in robotic research motivated by the generalization ability to novel objects~\cite{ReviewArticleSpringer_KBK}.
%
%
\mbox{Dex-Net}~\cite{Dex-Net_1.0,Dex-Net_2.0}
makes use of synthetic data
and locally analyzes the point cloud for finding robust parallel jaw grasps. The approach samples grasp candidates and ranks them by means of a neural network, which gets an aligned depth image crop and grasp candidate as input and outputs a grasp quality using a sigmoid output neuron. \mbox{Dex-Net 3.0}~\cite{Dex-Net_3.0} extends this framework to suction grippers and \mbox{Dex-Net 4.0}~\cite{Dex-Net_4.0} uses a multi-gripper policy, which automatically infers whether to use a parallel jaw or suction gripper for the next grasp execution. For the latter, two neural networks rank grasps
and an argmax policy chooses the highest quality grasp for execution.

Levine et al.~\cite{Google_2018_Learning_Hand-Eye_Coordination}
use 14 robots to execute 800,000 grasps over the course of two months and train a deep neural network for learning hand-eye coordination.
QT-Opt~\cite{QT_Opt} collects 560,000 grasp attempts with seven robots over several weeks and demonstrates
robust grasping and manipulation of a diverse set of objects based on reinforcement learning. Due to the immense amount of real-world data required, works such as GraspGAN~\cite{KonstantinosBousmalis.2018} and RCAN~\cite{RCAN} focus on reducing or avoiding the need for expensive and time-consuming data collection on real-world systems.

All approaches presented above solve pick-and-drop tasks and
do not propose a solution for precisely placing the picked candidates, even for rigid objects.

\begin{figure*}[t]
\begin{footnotesize}
\centering
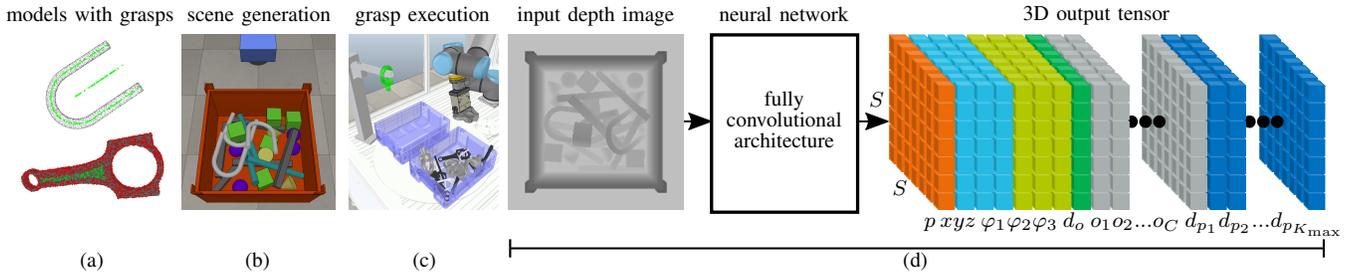
\caption{Overview of our approach: (a) 3D object models with automatically generated grasp poses for parallel jaw (a, top) and suction grippers (a, bottom).
(b) Physics simulation for scene generation by dropping random objects from a set of available objects into a bin.
(c) Physics simulation for grasping and object placement with a robot.
(d) The input of our model is a perspective depth image (d, left) that is processed by a fully convolutional network architecture (d, middle). The output of our neural network is a 3D tensor (d, right) comprising estimates for the presence of an object origin $\hat{p}$, positions $\hat{x}$, $\hat{y}$, $\hat{z}$, Euler angles $\hat{\varphi}_1$, $\hat{\varphi}_2$, $\hat{\varphi}_3$, relative pose distance for OPE $\hat{d}_o$, vector for the $C$ object classes, and the relative pose distance for object placement $\hat{d}_p$ for each grasp pose $\mat{G}_k \in \mathcal{G}$.}
\label{fig:Overview}
\end{footnotesize}
\vspace{-3mm}
\end{figure*}

\subsection{Model-based Approaches}
Analytic approaches typically match an object model to the sensor data~\cite{BinPicking_6D_OPE_IROS_2019,IPA_Diss_MatthiasPalzkill.2014}
and learning-based approaches map sensor data to pose estimates~\cite{OP-Net,OP-Net++,PPR_Net}. These outputs are then used to plan a collision-free and kinematically feasible grasp towards the object for picking~\cite{FPS_GPC_1,FPS_GPC_2,Spenrath.2013,Spenrath.2017}.
This is not sufficient for complex scenarios where jamming or entanglements~\cite{paper_MAM} with other objects in the bin can occur or objects move relative to the gripper because other objects are above them (examples see Fig.~\ref{fig:Motivation}).
%
Learning-based approaches allow to approximate these challenging and analytically difficult to describe correlations.
\mbox{PQ-Net} (Placement Quality Network)~\cite{PQ-Net} estimates object poses with graspability measures and
qualities for predefined grasps
based on grasping trials in simulation for a single object type.
The approach specifies a pose distance threshold and classifies grasp poses based on whether the grasp results in an object placement below the threshold or not, i.e., that all successful grasp poses are equally ranked.

In this work, we directly estimate the pose distance for object placement for different object types instead of specifying a distance threshold from the target pose and having multiple candidates which are regressed towards the same quality measure. This allows our extension of the Placement Quality Network (PQ-Net++) selecting the most promising grasp for a precise object placement on a global level.
Additionally, our approach estimates the pose distance from the ground truth
for the proposed pose estimates to select highly promising candidates for picking.
Furthermore, our approach autonomously decides which gripper to use
and thus allows maximizing the probability of a successful and precise object placement.

\section{Problem Statement}
\label{sec:ProblemStatement}
Given a robot kinematic as well as a discrete and finite set of grippers $\mathscr{E}$ with known gripper parameters (see Section~\ref{sec:GPG}), the task of the robot is to pick
known rigid objects from a
set of categories from a chaotic scene and place them at a given target pose $\mat{T} \in \mathrm{SE}(3)$.
%
Using a depth image $I$ of the scene from a single view, our goal is to localize objects, classify them, and identify a suitable grasp pose $\mat{G} \in \mathcal{G} \subset \mathrm{SE}(3)$ on the objects that allows a precise object placement.
The objects are localized by estimating their translation vector \mbox{$\mat{t}\in \NewR^3$} and rotation matrix \mbox{$\mat{R} \in \mathrm{SO}(3)$} relative to the sensor coordinate system.
Collisions of the manipulator with other objects in the scene have to be avoided. Additionally, picking objects that jam, get displaced relative to the gripper, or entangle with other objects in the scene have to be avoided.


\section{Placement Quality Network (PQ-Net++)}
\label{sec:Approach}
This section describes the techniques to automatically generate grasp poses for different gripper types, the synthetic data generation procedure, the required orientation unification step, the pose distance
definition, the parameterization of the output of the neural network, the multi-task loss function, the neural network architecture, the training procedure, the technique for a robust sim-to-real transfer, and the policy to infer robust grasps from the neural network output. Fig.~\ref{fig:Overview} illustrates an overview of our approach.

\subsection{Automatic Grasp Pose Generation}
\label{sec:GPG}
To avoid the need for manually specifying grasp poses $\mat{G}$
on the objects, we employ routines to automatically generate them. Based on a given gripper and object, the goal is to identify a discrete set of suitable grasp poses $\mathcal{G}$ on the object. The grasps $\mat{G} \in \mathcal{G}$ are defined relative to the object coordinate system.
In this paper, we focus on parallel jaw and suction grippers. For the former the gripper stroke as well as closing force and for the latter the suction cup dimensions as well as force and moment limits need to be known.

\begin{figure}[b]
\centering
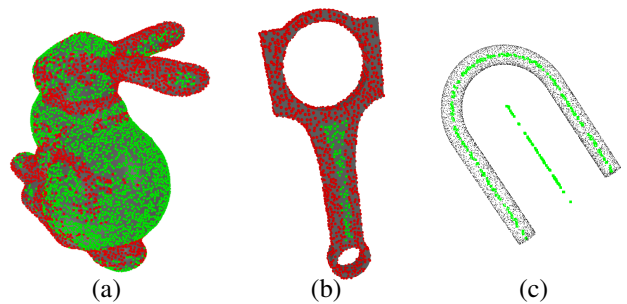
\caption{Automatically generated grasp poses exemplarily visualized for the (a) Stanford Bunny and (b) IPAConnectingRod~\cite{paper_MAM} for a suction gripper and (c) for the IPAUBolt for a parallel jaw gripper. Unsuitable grasp poses are shown in red and feasible grasp poses are highlighted in green.}
\label{fig:GraspPoses}
\end{figure}

To generate suction grasp candidates, points are sampled on the surface of the object. A grasp candidate consists of its position and the surface normal of the object at that position. Those candidates are then tested for a good seal formation by using a quasi-static spring model of a suction cup~\cite{Dex-Net_3.0,MA_Schnitzler}. If the grasp point is considered feasible, it is tested in simulation against collision of the gripper with the object.
For parallel jaw grippers, we use an automatic routine that samples grasp poses on the object and filters them based on the gripper stroke, normal information, and a simple collision check~\cite{Kleeberger_GPG_PJ,PQ-Net}.
Fig.~\ref{fig:GraspPoses} exemplarily shows automatically generated grasp poses for parallel jaw and suction grippers by our routines.

\subsection{Physics Simulation for Synthetic Data Generation}
For data generation, we use the physics simulation CoppeliaSim (formerly V-REP)~\cite{V-REP} with the built-in Bullet
physics engine to chaotically fill bins and execute grasping and placement trials in these scenes.

\subsubsection{Scene Generation}
Similar to the Sil\'{e}ane~\cite{Bregier_SymmetryAwareEvaluation} and Fraunhofer IPA~\cite{FraunhoferIPABinPickingDataset} datasets, we drop objects in random poses into a bin to generate chaotic scenes typical for \mbox{bin-picking}.
The object type to drop is randomly selected from the set of available objects.
The simulation records a perspective depth image of the scene together with the ID, class, pose relative to the sensor, and visibility \mbox{$v \in [0,1]$} of all objects in the scene.
The number of objects dropped in each scene is increased until a certain drop limit is reached, forming one cycle. This procedure gives a uniform distribution over different fill levels of the bin with uncorrelated images as the bins are completely refilled for each scene.

\subsubsection{Grasping and Placement}
In each generated scene, we loop over all grasp poses for each gripper for each object and check the collision-free reachability and kinematic feasibility of the grasp pose with the given gripper and robot kinematic. If the grasp pose can be reached without collisions, we execute the grasp and place the object at a defined target pose $\mat{T}$. Afterwards, we log the pose distance $d_{p} \in \NewR^+_0$ (see Section~\ref{sec:PoseDistance}) between the actual object placement and defined target pose while taking the proper symmetry class~\cite{Bregier_PoseDistance,Bregier_SymmetryAwareEvaluation} of the picked object into account. We also log if an entanglement occurred, i.e., that the lifted object is in contact with other objects from the scene as exemplarily visualized in Fig.~\ref{fig:Motivation}~(b).

The grasping of the objects is physically simulated, i.e.,
the object to pick is not simply rigidly attached to the gripper. For instance, when other objects lie on top of the targeted object, the picked candidate may become displaced relative to the parallel jaw gripper
as exemplarily illustrated in Fig.~\ref{fig:Motivation}~(a) or simply peel off
when using the suction gripper when the maximum pull force, shear force, or peel torque are exceeded.

\subsection{Orientation Unification}

Since the grasp poses $\mat{G}$
are defined relative to the object coordinate system, the orientation of the object has to be unified during data generation for grasping and placement to allow a proper learning of the correlations for the grasp poses and to avoid convergence issues with the grasp poses correlations during model training because of pose ambiguities due to object symmetries.
%
The orientation unification step also has to be applied at test time.

In this work, we assume that the $z$-axis of the object is the axis of symmetry.
However, it can be any other axis as long as it is known.
The orientation of all objects in the scene is unified relative to the camera coordinate system as follows depending on the proper symmetry class~\cite{Bregier_PoseDistance,Bregier_SymmetryAwareEvaluation} of the object.
%
For objects with no proper symmetry (e.g., Stanford bunny from Fig.~\ref{fig:GraspPoses}~(a)), no unification has to be applied as the orientation is already unique.
%
For spherical symmetries, we set $\mat{R}$ to the identity matrix $\mat{I}$.
%
For objects with a revolution symmetry (e.g., cone), we apply a rotation around the $z$-axis of the object coordinate system which minimizes the $z$-component of the $y$-axis. For revolution objects with rotoreflection invariance (e.g., cylinder), we additionally rotate around the $y$-axis by \ang{180} if the $x$-value of the $z$-axis can be increased.
%
For objects with finite symmetries around the $z$-axis only (e.g., IPAConnectingRod and IPAUBolt from Fig.~\ref{fig:GraspPoses} (b) and (c), respectively), we set the orientation with minimal $z$-value of the $y$-axis.
%
For objects with discrete symmetries around multiple axes (e.g., bar or cube), we choose the rotation matrix $\mat{R}\mat{H}_i$
with the smallest angle
\begin{equation}
\alpha=
\min\limits_{
\mat{H}_i \in \mathscr{H}
}\arccos\Big(\big(\Tr(\mat{R}\mat{H}_i)-1\big) \cdot 0.5\Big)
\end{equation}
of the axis–angle representation of a rotation matrix to the camera coordinate system.
$\mathscr{H} \subset \mathrm{SO}(3)$ is the discrete and finite set
of rigid transformations $\mat{H}_i$ that have no effect on the static state
of the object.

\subsection{Pose Distance}
\label{sec:PoseDistance}
Romain Br\'{e}gier et al.~\cite{Bregier_PoseDistance,Bregier_SymmetryAwareEvaluation} introduced a pose distance that allows an efficient distance computation for all possible kinds of object symmetries by introducing pose representatives $\mathscr{R}$ for all proper symmetry classes.
%
The representatives $\mathscr{R(P)}$ of pose
$\mathscr{P}=(\mat{R},\mat{t})$
depend on the proper symmetry class of the object.

Utilizing this representation framework, the distance $\Tilde{d} \in \NewR^{+}_0$ between a pair of poses $\mathscr{P}_1$, $\mathscr{P}_2$ can be expressed as the minimum Euclidean distance between their respective pose representatives:
\begin{equation}
\Tilde{d}(\mathscr{R}(\mathscr{P}_1),\mathscr{R}(\mathscr{P}_2)) =
\min\limits_{
\mat{p}_1 \in \mathscr{R}(\mathscr{P}_1),
\mat{p}_2 \in \mathscr{R}(\mathscr{P}_2)
}
\lVert \mat{p}_1-\mat{p}_2 \rVert
\label{eq:PoseDistance}
\end{equation}
This pose distance is compared against the commonly used threshold of 0.1 times the diameter $D \in \NewR^{+}$ of the smallest bounding sphere of the object $\mathcal{O}$
\begin{equation}
\Tilde{d} < 0.1 \cdot D
\end{equation}
to identify whether a pose estimate is considered as true or false positive~\cite{Bregier_SymmetryAwareEvaluation,Hinterstoisser_ADD/ADI_LINEMOD+}.

In this work, we consider the pose distance relative to 0.1 times the object diameter $D$ to 
deal with objects of different size
and, therefore, consider the relative pose distance
\begin{equation}
d=\frac{\Tilde{d}}{0.1 \cdot D}.
\end{equation}
If $d \in \NewR^{+}_0$ is smaller than 1, the distance $\Tilde{d}$ is below the distance threshold and the object pose estimate or object placement is considered as correct or precise enough~\cite{Bregier_SymmetryAwareEvaluation,PQ-Net}.

\subsection{Relative Pose Distance Estimation for Object Pose Estimation}
Our model estimates the relative pose distance $\hat{d}_o \in \NewR^+_0$ between the predicted and ground truth pose of the OPE for each predicted object pose while taking the proper symmetry class 
of the currently considered object
into account.
%
The ground truth
pose distance $d_o$ is determined based on the pose annotation from simulation and the pose output of the neural network and is, therefore, computed on the fly (dynamically during training), i.e., that the ground truth changes during training.

\subsection{Parameterization of the Output}
Similar to~\cite{OP-Net,OP-Net++} and~\cite{PQ-Net}, we employ a spatial discretization to split the measurement volume of the sensor into $S \times S$ volume elements. Each object has a $(8+C+K_\mathrm{max})$-dimensional feature vector $\mat{x}$ comprising
the presence $p=1$ of an object origin, the positions $x$, $y$, $z$ and Euler angles $\varphi_1$, $\varphi_2$, $\varphi_3$ relative to the camera coordinate system, the relative pose distance $d_o$ of the proposed object pose estimate (ground truth is computed on the fly), the $C$-dimensional one-hot encoded object class vector, and the $K_\mathrm{max}$-dimensional vector with the relative pose distance for object placement for each grasp pose $\mat{G}_k$ as defined for the given object class
where $K_\mathrm{max}$ is the maximum number of grasp poses from all objects.

The ground truth output tensor of the neural network is initialized with zeros. The feature vectors of the objects are assigned to the tensor entries, based on the position of the object origin in the scene. In case multiple objects fall into the same spatial location, the feature vector $\mat{x}$ of the object with higher visibility $v$ is used as ground truth.
The $S \times S \times (8+C+K_\mathrm{max})$ output tensor is visualized in Fig.~\ref{fig:Overview}~(d, right).

To generate the ground truth vector with the object placement annotations, we
concatenate the vectors for different grippers.
%
As the objects $\mathcal{O}_j$ with $j=1,...,C$
generally do
not have the same number of grasp poses $K_j$, we simply fill the remaining entries in the $K_\mathrm{max}$-dimensional vector with the default value $d_\mathrm{max}$ if $K_j < K_\mathrm{max}$.
%
If an entanglement has occurred, we set $d_{p_k}=d_\mathrm{max}$ for that grasp pose $\mat{G}_k$ to prevent our policy from selecting entangled grasps. Furthermore, we clip the relative pose distance for object placement $d_p$ at $d_\mathrm{max}$ for all executed grasp poses because the pose distance is unpredictably high for dropped objects. Moreover, we set the ground truth pose distance to $d_\mathrm{max}$ for the grasp poses which were not executed (due to no collision-free reachability and/or kinematic feasibility).
In our experiments, we use $d_\mathrm{max}=1$.
%
At test time, we only consider the first $K_j$ entries in the relative pose distance
vector for object placement based on the object class prediction.

\subsection{Loss Function}

To train the neural network, the multi-task loss function
\begin{equation}
\mathcal{L} = \sum_{i=1}^{S^2} \bigg( \lambda_1\mathcal{L}_p+\Big[ \lambda_2\mathcal{L}_\mathrm{pose}+
\lambda_3\mathcal{L}_{d_o}+
\lambda_4\mathcal{L}_\mathrm{cls}+
\lambda_5\mathcal{L}_{d_p} \Big] \lambda_6 p_i \bigg)
\end{equation}
with manually tuned weights $\lambda$ is optimized. In our experiments, we use $\lambda_1=0.1$, $\lambda_2=0.5$, $\lambda_3=0.01$, $\lambda_4=0.01$, $\lambda_5=0.5$, and $\lambda_6=8v^3$. The weighting with the ground truth visibility $v$ causes the neural network to prioritize the more visible objects.

The loss term $\mathcal{L}_p$ reflects the presence of object origins in the volume elements and is defined via the binary cross-entropy.
For the pose loss $\mathcal{L}_\mathrm{pose}$, we use Equation~(\ref{eq:PoseDistance}), which properly considers all possible kinds of object symmetries.
For the object classification loss $\mathcal{L}_\mathrm{cls}$, we use the 
categorical cross-entropy loss.
For the loss of the relative pose distance for OPE (dynamic ground truth) and object placement for each grasp pose (static ground truth) with $\mathcal{L}_{d_o}$ and $\mathcal{L}_{d_p}$, respectively, we use the squared L2 norm.

\subsection{Neural Network Architecture}
Our model gets a single normalized perspective depth image $I$ with depth values $d_s \in [-1,1]$ bounded by the near and far clipping plane of the sensor as input and outputs the $S \times S \times (8+C+K_\mathrm{max})$ tensor as visualized in Fig.~\ref{fig:Overview}~(d). In our experiments, we use a DenseNet-BC~\cite{DenseNet,DenseNet_JournalPaper} with 40 layers and a growth rate of 50 as function approximator for the mapping from inputs to outputs. The growth rate specifies the number of feature maps being added per layer within a dense block.
We use a $128 \times 128$ input depth image and a $16\times16\times1$ spatial discretization for the $x$, $y$, and $z$ direction, respectively, while downsampling is performed via average pooling.
%
ReLU activation functions are employed in the hidden layers. For the 3D output tensor, we use sigmoid functions for the presence of object origins and the relative pose distance estimation for object placement, linear functions for the pose information channels, ReLU functions for the relative pose distance estimation for OPE, and vector-wise softmax functions for object classification.
With this network architecture, forward passes are performed with an average frame rate of 92~fps or 44~fps on a Nvidia Tesla V100 or a GTX 1080 Ti, respectively.

\subsection{Training}
During training, the loss of the entire channel for object origin presence estimation is backpropagated. For all other channels, we only backpropagate the error of the entries in the tensor with assigned feature vector $\mat{x}$ data by multiplying channel-wise with the ground truth object origin presence channel.
Due to the time-consuming process of grasping and placement, not all generated samples obtain annotations for the grasp poses and we only backpropagate $\mathcal{L}_{d_p}$ if annotations are given.


\begin{figure}[tb]
\centering
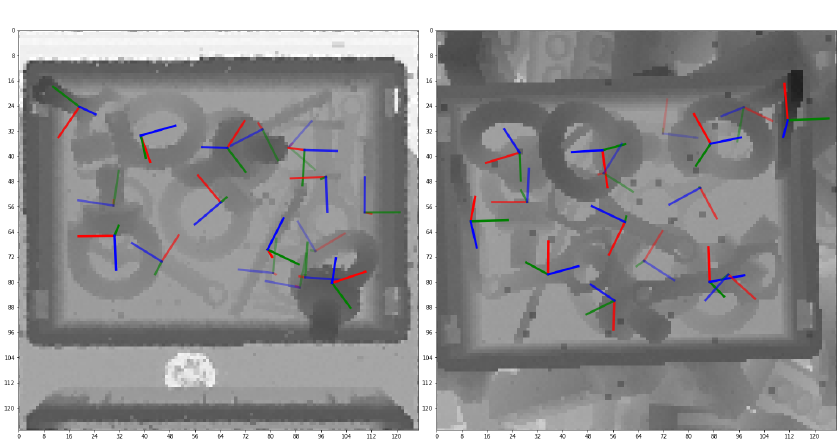
\caption{Exemplary augmented training samples for transferring our model from simulation to the real world.}
\label{fig:samples_augmented}
\vspace{-3mm}
\end{figure}

\subsection{Sim-to-Real Transfer}
We use the sim-to-real transfer technique domain randomization~\cite{Domain_Randomization} to successfully deploy our model from simulation to the real world. During scene generation, we randomize the bin pose as well as the position and orientation of the objects above the bin before dropping.

During training, we insert a random background image from the ``\textit{Fraunhofer IPA Bin-Picking dataset}''~\cite{FraunhoferIPABinPickingDataset} with a probability of 0.5 (see Fig.~\ref{fig:samples_augmented}, right). Otherwise, the default background from the setup is used (see Fig.~\ref{fig:samples_augmented}, left). Additionally, we apply image augmentations such as blurring, elastic transformations, and adding noise to the synthetic depth images. As real-world data may also obtain wrong and not only noisy data due to, e.g., reflections for shiny metal parts (as it is the case for the IPAUBolt and IPARingScrew; see Fig.~\ref{fig:result_real_world_data}, left),
we change the value of random single pixels and larger regions in the image to an increased depth value above the actual object (drop noise). All image augmentations are applied with varying intensity and in a random order during training.
Fig.~\ref{fig:samples_augmented} exemplarily visualizes images used for training while the flawless object annotations
from simulation are kept.

\subsection{Policy}


The deep neural network $f$ with trained weights $\theta$ gets a single normalized perspective depth image $I$ with global scene information as input and outputs a 3D tensor $\hat{T}$.
%
%
The function $g(\hat{T})=\Tilde{T}$ transforms the relative pose distance estimate for OPE $\hat{d}_o$ and the relative pose distance estimates for object placement $\hat{d}_{p_k}$ for each grasp pose $\mat{G}_k$ in the output tensor $\hat{T}$ to a score
with
\begin{equation}
s_{o,i} = \exp(-0.5 \cdot \hat{d}_{o,i})
\end{equation}
and
\begin{equation}
s_{p_k,i}=1-\hat{d}_{p_k,i}/d_\mathrm{max}.
\end{equation}
%
Since $\hat{d}_{p_k}$ is bounded by $d_\mathrm{max}$, we do not use an exponential function.
Our policy
\begin{equation}
\pi\big(g(f_\theta(I))\big) = \argmax_{i,k}(
\hat{p}_i \cdot
s_{o,i}
\cdot s_{p_k,i})
\end{equation}
selects the highest quality object with grasp pose $\mat{G}_k$ from all $S^2$ volume elements for execution with $i=1,...,S^2$ and $k=1,...,K_j$ with $K_j$ being the number of predefined grasp poses for object $\mathcal{O}_j$ (predicted object class) for all
grippers. For each grasp pose $\mat{G}_k$, the corresponding gripper is known.

\section{Experiments}
\label{sec:Experiments}


For performance evaluation, we introduce two novel synthetic benchmark datasets: Mixed bins \textit{symmetries} and \textit{entanglements}. Both datasets comprise objects from multiple categories and of different size and shape.

The \textit{symmetries} dataset comprises seven symmetric objects from all possible proper symmetries classes (see Fig.~\ref{fig:Overview} (b) and (d, left) as well as Fig.~\ref{fig:Motivation}~(a), (c), and (d)). Using this dataset, we demonstrate that our method works for all possible kinds of object symmetries.
%
Furthermore, the dataset comprises very elongated objects (cylinder and bar). When attempting to pick these objects, they can easily move relative to the gripper. Thus, a proper picking order of the objects has to be considered and simply checking collision-free reachability of the grasp poses is not sufficient.

The \textit{entanglements} dataset is very challenging because the three objects can potentially entangle (see Fig.~\ref{fig:Motivation}~(b)), i.e., that no precise placement is possible anymore due to collisions of the additionally picked object with the environment. Exemplary images from this dataset are visualized in Fig.~\ref{fig:samples_augmented} and the application to real-world data in Fig.~\ref{fig:result_real_world_data}.

In the following, we benchmark the performance of our approach for OPE, grasping, and precise object placement in simulation on the challenging mixed bins \textit{symmetries} and \textit{entanglements} datasets.
Since the baseline approaches \mbox{OP-Net}~\cite{OP-Net} and \mbox{PQ-Net}~\cite{PQ-Net} are not originally designed to handle different object classes, we also extended them to object classification by adding $C$ output feature maps and backpropagating $\mathcal{L}_\mathrm{cls}$ during training while using the same network architecture as for our approach.
%
%
All performance numbers are recorded on test datasets of 10 cycles each (500 samples).
For the experiments in Section~\ref{Sec:BenchmarkingGrasping} and~\ref{Sec:BenchmarkingObjectPlacement}, each approach executes exactly one grasp per scene. This allows a fair comparison because all approaches observe the exact same scenarios.
%

\subsection{Benchmarking 6D Object Pose Estimation} 
\mbox{PQ-Net}~\cite{PQ-Net} used grasping and placement annotations for around 13\% of the cycles only and reported an average precision (AP) for OPE slightly worse than \mbox{OP-Net}~\cite{OP-Net} across all datasets. In our datasets, we annotated 50\% of the samples with grasps.
Table~\ref{table_benchmarking_OPE} reports the AP values based on the metric from Br\'{e}gier et al.~\cite{Bregier_PoseDistance,Bregier_SymmetryAwareEvaluation} for the two challenging mixed bins datasets.
The AP values are determined independently for each object class and averaged afterwards.
With enough training samples with grasping annotations, \mbox{PQ-Net++} is a
better pose estimator than \mbox{OP-Net}~\cite{OP-Net}.
Adding additional outputs has positive effects on OPE due to multi-task learning, even if the output size of the neural network increases a lot by adding a feature map for each grasp pose $\mat{G}_k$.
Additionally, we report very high success rates for object classification for the objects with ground truth visibility $v\geq0.5$.

\begin{table}[tbp]$ $
\begin{tiny}
\caption{Performance evaluation for 6D object pose estimation and object classification.
Higher is better.
}
\label{table_benchmarking_OPE}
\begin{center}
\begin{tabular}{|l||c|c|}
\hline
& mixed bins & mixed bins \\
& symmetries dataset & entanglements dataset \\
\hline
\hline
AP average \mbox{OP-Net}~\cite{OP-Net} extended to object classification & 0.80 & 0.81 \\
\hline
AP average \mbox{PQ-Net++} (ours) & \textbf{0.88} & \textbf{0.83} \\
\hline
\hline
success rate object classification \mbox{OP-Net}~\cite{OP-Net} & \textbf{0.99} & \textbf{0.98} \\
\hline
success rate object classification \mbox{PQ-Net++} (ours) & \textbf{0.99} & \textbf{0.98} \\
\hline
\end{tabular}
\end{center}
\end{tiny}
\vspace{-3mm}
\end{table}

\begin{table}[b]
\begin{tiny}
\caption{Success rate for grasping. Higher is better.
}
\label{table_benchmarking_grasping}
\begin{center}
\begin{tabular}{|l||c|c|}
\hline
& mixed bins & mixed bins \\
& symmetries dataset & entanglements dataset \\
\hline
success rate grasping \mbox{Dex-Net 4.0}~\cite{Dex-Net_4.0} & 0.77 & 0.59 \\
\hline
success rate grasping \mbox{PQ-Net++} (ours) & \textbf{0.98} & \textbf{0.94} \\
\hline
\end{tabular}
\end{center}
\end{tiny}
\vspace{-3mm}
\end{table}

\begin{table*}[htbp]$ $
\begin{tiny}
\caption{
Results for object placement:
Success rate (higher is better) and average relative pose distance for object placement (at a defined target pose) (lower is better).
Evaluation of the 6D object pose estimation (OPE) of the chosen objects for picking:
Success rate for OPE based on commonly used evaluation metrics (higher is better) and average relative pose distance for OPE (lower is better).
Best results are marked in bold.
}
\label{table_benchmarking_placement}
\begin{center}
\begin{tabular}{|l||c|c|}
\hline
& mixed bins & mixed bins \\
& symmetries dataset & entanglements dataset \\
\hline
\hline
success rate object placement \mbox{PQ-Net}~\cite{PQ-Net} extended to object classification and multiple grippers & \textbf{0.96} & 0.82 \\ 
\hline
success rate object placement \mbox{PQ-Net++} ($\hat{p}$, $s_p$) (ours) & \textbf{0.96} & \textbf{0.83} \\ 
\hline
success rate object placement \mbox{PQ-Net++} ($\hat{p}$, $s_o$, $s_p$) (ours) & 0.94 & 0.82 \\ 
\hline
\hline
avg. relative pose distance object placement \mbox{PQ-Net}~\cite{PQ-Net} extended to object classification and multiple grippers & 0.86 & 0.47 \\ 
\hline
avg. relative pose distance object placement \mbox{PQ-Net++} ($\hat{p}$, $s_p$) (ours) & \textbf{0.16} & \textbf{0.35} \\ 
\hline
avg. relative pose distance object placement \mbox{PQ-Net++} ($\hat{p}$, $s_o$, $s_p$) (ours) & 0.19 & 0.36 \\ 
\hline
\hline
success rate OPE based on~\cite{Bregier_PoseDistance,Bregier_SymmetryAwareEvaluation} metric \mbox{PQ-Net}~\cite{PQ-Net} extended to object classification and multiple grippers & \textbf{0.99} & 0.89 \\ 
\hline
success rate OPE based on~\cite{Bregier_PoseDistance,Bregier_SymmetryAwareEvaluation} metric \mbox{PQ-Net++} ($\hat{p}$, $s_p$) (ours) & \textbf{0.99} & 0.87 \\ 
\hline
success rate OPE based on~\cite{Bregier_PoseDistance,Bregier_SymmetryAwareEvaluation} metric \mbox{PQ-Net++} ($\hat{p}$, $s_o$, $s_p$) (ours) & \textbf{0.99} & \textbf{0.90} \\ 
\hline
\hline
success rate OPE based on ADI metric~\cite{Hinterstoisser_ADD/ADI_LINEMOD+} \mbox{PQ-Net}~\cite{PQ-Net} extended to object classification and multiple grippers & \textbf{1.0} & 0.98 \\ 
\hline
success rate OPE based on ADI metric~\cite{Hinterstoisser_ADD/ADI_LINEMOD+} \mbox{PQ-Net++} ($\hat{p}$, $s_p$) (ours) & \textbf{1.0} & \textbf{0.99} \\ 
\hline
success rate OPE based on ADI metric~\cite{Hinterstoisser_ADD/ADI_LINEMOD+} \mbox{PQ-Net++} ($\hat{p}$, $s_o$, $s_p$) (ours) & \textbf{1.0} & \textbf{0.99} \\ 
\hline
\hline
avg. relative pose distance OPE \mbox{PQ-Net}~\cite{PQ-Net} extended to object classification and multiple grippers & 0.18 & 0.59 \\ 
\hline
avg. relative pose distance OPE \mbox{PQ-Net++} ($\hat{p}$, $s_p$) (ours) & 0.15 & 0.59 \\ 
\hline
avg. relative pose distance OPE \mbox{PQ-Net++} ($\hat{p}$, $s_o$, $s_p$) (ours) & \textbf{0.13} & \textbf{0.56} \\ 
\hline
\end{tabular}
\end{center}
\end{tiny}
\vspace{-3mm}
\end{table*}

\subsection{Benchmarking Grasping}
\label{Sec:BenchmarkingGrasping}

Since, to the best of our knowledge, there are no other model-based approaches explicitly designed for mixed bins with multiple grippers, we compare our approach with \mbox{Dex-Net 4.0}~\cite{Dex-Net_4.0} because it also uses multi-gripper policies to pick objects from chaotic bins.
Table~\ref{table_benchmarking_grasping} reports the success rates for grasping in simulation. A grasp is considered as successful if an object is in the gripper after moving out of the bin.
\mbox{Dex-Net} is
not explicitly trained on the considered objects.
The approach analyzes the point cloud locally and does contrary to our approach not reason on a global level. Thus, it does not explicitly attempt grasps close to the center of mass for very elongated objects (e.g., bar and cylinder) or not occluded objects.
Moreover, \mbox{Dex-Net} operates in 4D only (top-down grasps) for the parallel jaw gripper, which may result in less stable grasps, whereas our approach operates in 6D.
Our approach performs better because it uses model knowledge and is explicitly trained on the scenarios. Our experiments demonstrate that the system performance can be improved by
incorporating model knowledge.



\subsection{Benchmarking Object Placement} 
\label{Sec:BenchmarkingObjectPlacement}



\mbox{PQ-Net}~\cite{PQ-Net} specifies a distance threshold of 0.1 times the object diameter $D$ for object placement and performs a binary classification for each grasp pose based on whether the placement trial resulted in a placement within the threshold or not, i.e., that all grasp poses within the threshold are treated equally.
%
%
We compare this baseline with our policy with and without the relative pose distance estimation for OPE.

Table~\ref{table_benchmarking_placement} reports the results of our object placement experiments in simulation.
The object placement success rate identifies whether the placement is within the distance threshold for all samples, where both approaches perform very similar.
In our experiments, we log the relative pose distance for object placement for each
trial and average it over all samples from the test dataset.
%
As we directly estimate the pose distance for object placement for each grasp pose, we are able to select grasp poses for more precise object placements. Thus, our approach chooses grasp poses that result in significantly more
precise object placements than \mbox{PQ-Net}.




Additionally, we evaluate the correctness of the OPE of the chosen candidate from the policy (without ICP refinement).
Table~\ref{table_benchmarking_placement} reports the success rates for OPE based on the metric from Br\'{e}gier et al.~\cite{Bregier_PoseDistance,Bregier_SymmetryAwareEvaluation} and the ADI~\cite{Hinterstoisser_ADD/ADI_LINEMOD+} metric. For both metrics very high success rates are reported.
Furthermore, we determine the average relative pose distance for OPE over all samples. Our policy with the pose distance estimate $\hat{d}_o$ provides a better results for OPE because of explicitly focusing on more correct pose estimates.

Our method learns to avoid the failure cases presented in Fig.~\ref{fig:Motivation}~(a) and (b) and can also pick objects
close to the bin wall
(see Fig.~\ref{fig:Motivation}~(c)) or very close together (see Fig.~\ref{fig:Motivation}~(d)) due to employing grasps from multiple grippers.


\section{Conclusions}
\label{sec:Conclusions}
In this paper, we introduced a novel approach for grasping and placing known rigid objects of different categories from highly cluttered scenes
using multi-gripper policies.
Based on a depth image of the scene,
%
our approach estimates the 6D object pose together with a pose distance estimate for object pose estimation and a quality estimate for each automatically generated and predefined grasp pose on the object for multiple objects simultaneously in a single forward pass of the network.

Due to incorporating model knowledge into the system, our approach reports higher success rates for grasping than state-of-the-art model-free systems and additionally allows precisely placing the picked objects. By estimating the pose distance for object placement for each grasp pose, our approach selects the most promising grasp for a precise object placement on a global level and allows placing objects significantly more precise than prior work.
%
%
%
Our experiments demonstrate that our approach scales to a high level of complexity with operating for mixed bins, all possible kinds of object symmetries, and multiple grippers in combination.


\addtolength{\textheight}{-1cm}  



\section*{Acknowledgment}
This work was partially supported by the
Federal Ministry of Education and Research (Deep Picking -- Grant No. 01IS20005C), the
State Ministry of Baden-W\"urttemberg for Economic Affairs, Labour and Housing Construction
(Center for Cognitive Robotics –- Grant No. 017-180004 and Center for Cyber Cognitive Intelligence (CCI) -- Grant No. 017-192996), and the Fraunhofer lighthouse project SWAP.
We would like to thank
our colleagues for helpful discussions
and Lucas Doust Alba for the support with the \mbox{Dex-Net} experiments.


\bibliographystyle{IEEEtran}
\bibliography{IEEEabrv,references}

\end{document}